\documentclass[11pt,a4paper]{article}
\usepackage{fontspec}
\usepackage[english]{babel}
\usepackage{fix-cm}
\usepackage[top=3cm,bottom=3cm,left=3cm,right=2cm,headheight=13pt,headsep=28pt,footskip=45bp]{geometry}
\usepackage{amsmath,amssymb,graphicx,booktabs,tabularx,array}
\usepackage{caption,titlesec,fancyhdr,enumitem,float}
\usepackage{fvextra}
\usepackage{csquotes}
\usepackage{xurl}
\usepackage[style=apa,backend=biber,sortcites=true,doi=true,url=false,isbn=false]{biblatex}
\usepackage[hidelinks,pdfauthor={Thilo Tamme; Anton Hantel; Jakob Schepers},pdftitle={Who Writes the Codebook? An Eight-Step Framework for Auditing LLM-Generated Category Systems}]{hyperref}
\AtBeginDocument{\fontsize{11bp}{12.65bp}\selectfont}
\titleformat{\section}{\sffamily\bfseries\fontsize{14bp}{16bp}\selectfont}{\thesection}{1em}{}
\titleformat{\subsection}{\sffamily\bfseries\fontsize{12bp}{14bp}\selectfont}{\thesubsection}{0.8em}{}
\titlespacing*{\section}{0pt}{12bp}{6bp}
\titlespacing*{\subsection}{0pt}{11bp}{6bp}
\DeclareCaptionFont{eciscaption}{\fontsize{11bp}{12.65bp}\selectfont}
\newcommand{\tablefont}{\fontsize{10bp}{11.5bp}\selectfont}
\newcommand{\tabnote}[1]{\par\vspace{3bp}{\tablefont #1\par}}
\newcolumntype{L}[1]{>{\raggedright\arraybackslash}p{#1}}
\newcolumntype{Y}{>{\raggedright\arraybackslash}X}
\setlist{leftmargin=1.5em,itemsep=2bp,topsep=3bp,parsep=0pt}
\fancypagestyle{firstpage}{\fancyhf{}\fancyfoot[L]{\fontsize{10bp}{12bp}\selectfont\itshape Preprint}\fancyfoot[R]{\fontsize{10bp}{12bp}\selectfont\thepage}}
\begin{document}
\thispagestyle{firstpage}
\begin{center}
{\bfseries\fontsize{16bp}{18.4bp}\selectfont WHO WRITES THE CODEBOOK? AN EIGHT-STEP FRAMEWORK FOR AUDITING LLM-GENERATED CATEGORY SYSTEMS\par}
\vspace{6.5bp}
{\itshape\fontsize{12bp}{14bp}\selectfont Preprint\par}
\end{center}
\vspace{12bp}
\begin{center}
{\fontsize{12bp}{14bp}\selectfont Thilo Tamme\textsuperscript{1,*}\quad Anton Hantel\textsuperscript{2,*}\quad Jakob Schepers\textsuperscript{1}\par}
\vspace{8bp}
{\fontsize{11bp}{12.65bp}\selectfont
\textsuperscript{1}Technical University of Munich (TUM)\par
\textsuperscript{2}Massachusetts Institute of Technology (MIT)\par}
\vspace{3bp}
{\fontsize{10bp}{11.5bp}\selectfont\textsuperscript{*}These authors contributed equally to this work.\par}
\end{center}
\vspace{6bp}
\section*{Abstract}
{\itshape Large language models (LLMs) increasingly create the categories through which qualitative evidence is interpreted. Existing coding benchmarks often hold these categories fixed. We propose the Qualitative Variance Audit Framework, an eight-step method that compares codebook creation, repeated application and code distributions under controlled inputs. It combines model-owned and shared-reference codebooks to distinguish variation in the overall pipeline from application differences within a common category system. An application to 668 organisational interviews, 13,166 text units and eight models from five vendors yields 48 codebooks and 270 coding runs. Models create markedly different category structures; mean within-model reliability ranges from .76 to .92, whereas between-model agreement on the shared codebook is about .69. The framework complements LLM-assisted grounded theory by exposing model dependencies before substantive interpretation. It assesses reliability and representational consequences, not interpretive validity or the quality of resulting theory.\par}
\vspace{3bp}
{\itshape Keywords: Generative AI, Qualitative methods, Codebook development, Intercoder reliability, Qualitative variance.\par}

\section{Introduction}
Language models increasingly mediate both the collection and the analysis of qualitative evidence. In organisational research, this makes model choice consequential twice: an AI interviewer can shape what respondents are asked, and an AI analyst can determine which distinctions their answers contain. Information systems (IS) research offers methods for taxonomy development \parencite{nickerson2013method} and computational theory development \parencite{berente2019}. The next-generation methods agenda in the \emph{Journal of Information Technology} explicitly asks how generative AI changes research practice and how its contributions should be evaluated \parencite{blohm2025}. A specific methodological problem remains: how can researchers diagnose dependence on the model that creates their qualitative coding instrument?

Codebook-based analysis involves both defining categories and assigning material to them. Deductive coding studies evaluate assignment under a supplied codebook \parencite{xiao2023deductive,tai2024}. Inductive studies already investigate generated categories, repeated runs and comparisons with human analysis \parencite{depaoli2024inductive,arlinghaus2024,dorfel2026}. What is needed is a reusable diagnostic design linking the structure of a generated codebook to the reliability and distribution of its subsequent application. A model can repeatedly apply a category system consistently while that system suppresses distinctions another model would preserve. Agreement alone does not reveal this representational choice.

We ask: \emph{How can researchers audit model dependence in LLM-generated category systems and their application to a common qualitative corpus?} We answer with the \emph{Qualitative Variance Audit Framework} (QVAF), organised into eight operational steps. It fixes the elicitation sample and coding units, obtains model-specific codebooks, applies model-owned and shared-reference codebooks repeatedly, and reports structural, reliability and distributional diagnostics. The shared condition compares application within one common category system; the own-codebook condition describes the combined consequences of model and instrument choice. This is a diagnostic comparison rather than a causal decomposition of independent model and codebook effects.

The methodological contribution is the integration of these controls and diagnostics into a transferable audit procedure. It is not the first use of LLMs for inductive coding, nor a new grounded-theory method. Grounded-theory workflows use coding to develop conceptual explanations; QVAF examines model dependencies in coding instruments and outputs that may feed that work. A second contribution is an empirical demonstration across eight models from five vendors, six themes and 13,166 units from 668 AI-led interviews. A third is a bounded interpretation of \emph{qualitative variance}: the spread of code assignments conditional on a stated codebook and coder panel, rather than an intrinsic property of the corpus.

The application yields 48 codebooks and 270 coding runs. Mean codebook size varies from 15 to 25 subcodes, and models disagree on which codes admit ordered levels. Mean within-model reliability on own codebooks ranges from .76 to .92, while between-model agreement on the shared codebook is about .69. These findings motivate cross-model sensitivity checks before coded distributions are treated as substantive evidence. No human reference analysis was conducted, so the study does not identify the best codebook or establish validity.

\section{Related Work and Methodological Positioning}
\subsection{From coding assistance to generated category systems}
Work on LLM-assisted qualitative analysis spans different analytic tasks. \textcite{xiao2023deductive} combine a supplied codebook with GPT-3 for deductive coding; \textcite{tai2024} examine repeated coding and comparison with human analysis. These designs test application under specified categories. Inductive work extends the task to category generation. \textcite{depaoli2024inductive} explores inductive thematic analysis of interviews, while TopicGPT provides a prompt-based framework for generating and assigning topics \parencite{pham2024topicgpt}. These contributions establish that categories themselves can be outputs of computational analysis.

The closest comparisons already go beyond a single demonstration. LAIC separates category formation from assignment and compares GPT models across temperatures and repeated runs \parencite{arlinghaus2024}. \textcite{bennis2025} compare thematic analyses from nine generative models. \textcite{dorfel2026} evaluate a generated code system and its application against human qualitative content analysis, including structural and semantic comparisons. QVAF contributes a specific combination: common induction samples and fixed units, repeated coding under own and shared-reference instruments, and linked reporting of category structure, coverage, agreement and code-use dispersion across vendors and themes.

\subsection{Relationship to grounded theory and IS methods}
LLM-supported grounded theory addresses a different inferential objective. \textcite{sinha2024} examine GPT-4 in exploratory and focused coding, including generation, comparison and refinement of codes within an interpretive research process. Generative grounded theory offers a seven-step process extending from corpus formation to conceptual abstraction, theoretical integration and saturation assessment \parencite{schmitt2026}. Human interpretation, traceability and a staged workflow are therefore established methodological features.

QVAF supplies an audit that can accompany such workflows. It asks whether category structures and coded evidence change when another model processes the same material. It does not conduct theoretical sampling, establish theoretical saturation, integrate categories into theory or assess the explanatory quality of that theory. Its output is an account of model dependence and issues for researcher inspection. In IS taxonomy development, it helps inspect candidate distinctions before accepting an AI-generated taxonomy; the audit does not replace the conceptual and empirical evaluation required by \textcite{nickerson2013method}. In computational theory development, it makes visible a choice in the transformation of text into analysable categories \parencite{berente2019}. This operational focus addresses transparency and evaluation concerns raised by \textcite{blohm2025}.

\subsection{Reliability and codebook-conditional variance}
QVAF is situated in the coding-reliability tradition, where explicit definitions and shared procedures permit comparison among coders \parencite{macqueen1998codebook,oconnor2020intercoder}. It is not a universal quality standard for qualitative research. Reflexive thematic analysis treats researcher interpretation differently and does not make intercoder agreement its quality criterion \parencite{braun2019reflecting}. The framework is appropriate when a study deliberately uses explicit codebooks and wants to assess the sensitivity of coded representations to model choice.

Reliability concerns consistency, not correctness \parencite{krippendorff2004reliability,hayes2007answering}. Repeated calls to one model share training and other priors, so agreement cannot substitute for independent validation. Systematic errors may remain even when coding is stable \parencite{ashwin2025bias}. Fixing text units is also necessary because agreement otherwise mixes coding and segmentation differences \parencite{campbell2013coding}.

For nominal code distributions, entropy and its effective-number transformation describe how assignments spread across categories \parencite{hill1973diversity,jost2006entropy}. We use \emph{qualitative variance} in this distributional sense. It does not denote a variance-theoretic causal model, statistical variance of a continuous construct or the conceptual richness of an interpretation. Its value depends on the available categories, their granularity and the coders applying them. Comparisons require an explicit reference codebook and a consistent counting rule.

\section{The Qualitative Variance Audit Framework}
\subsection{Eight steps and their outputs}
QVAF makes the benchmark reusable by specifying an input, operation and auditable output at each step (Table~\ref{tab:framework}). Steps 1--3 construct comparable instruments, Steps 4--5 apply them, and Steps 6--8 diagnose and interpret their consequences. The eight-step organisation formalises the workflow reported here; the empirical application demonstrates its feasibility and diagnostic value, not independent validation of every design choice.

\begin{table}[!htbp]
\tablefont
\begin{tabularx}{\textwidth}{L{3.0cm}YY}
\toprule
\textbf{Step} & \textbf{Operation and control} & \textbf{Auditable output}\\
\midrule
1. Fix the audit corpus & Define themes, exclusions and one segmentation rule; freeze unit identifiers. & Corpus inventory and identical coding units.\\
2. Match elicitation inputs & Select the same interview sample per theme for every model; hold prompt and schema constant. & Sample identifiers and creation specification.\\
3. Generate codebooks & Obtain labels, definitions, rules, examples and scale declarations; validate structure. & Model-specific instruments and structural diagnostics.\\
4. Set comparison conditions & Pair each model's own codebook with one shared reference selected by a declared rule. & Own/shared design and explicit reference dependence.\\
5. Repeat corpus coding & Apply each condition to all fixed units in separate runs; record settings and failures. & Repeated assignments and execution manifest.\\
6. Assess representation and agreement & Compare granularity, ordinal share, coverage, within-model reliability and between-model agreement. & A diagnostic profile rather than one quality score.\\
7. Measure conditional dispersion & Compute code-use distributions within the shared category system. & Entropy, effective code count and dominance, with denominators.\\
8. Interpret and document & Relate discrepancies to substantive decisions; distinguish evidence from unresolved validation. & Bounded conclusions and targets for human review.\\
\bottomrule
\end{tabularx}
\caption{The eight-step Qualitative Variance Audit Framework.}
\label{tab:framework}
\end{table}

\textbf{Step 1: fix the audit corpus.} Researchers define the analytic scope before comparing models. Deployments with different protocols or audiences remain separate themes. Cleaning and segmentation are performed once, and every coding condition receives the same unit identifiers and text. This prevents unit-boundary differences from being mistaken for interpretive differences. Exclusions and the availability of conversational context must be reported.

\textbf{Step 2: match elicitation inputs.} All models create a codebook from the same sampled interviews within each theme. The sample must fit the smallest context window. Common prompts and schemas constrain the task without forcing identical distinctions. The sample is an elicitation control, not evidence of conceptual saturation. The application uses 15 interviews per theme; reported saturation ranges can inform this choice \parencite{hennink2022sample}, but do not validate it for a particular corpus.

\textbf{Step 3: generate and validate codebooks.} Each model returns a two-level category system. Every subcode has a label, definition, inclusion and exclusion rules, example and nominal/ordinal declaration; ordinal codes also specify ordered levels. These components make the instrument inspectable and applicable by another coder \parencite{macqueen1998codebook}. Neutral identifiers are assigned after generation. A fixed residual code records relevant content that fits no substantive code. Structural validation enforces the schema and rejects additional catch-all categories; it does not establish meaningfulness or completeness.

\textbf{Step 4: set comparison conditions.} Each model applies both its own codebook and the shared-reference codebook. The first represents its complete creation-and-application pipeline. The second holds categories fixed for comparing application. Choosing one reference limits computation, but makes the comparison conditional on that reference. With $M$ models, $T$ themes and $R$ repeated coders, using one model's codebooks as the reference requires $TR(2M-1)$ coding runs; a fully crossed design requires $TRM^2$. The reference model's duplicate condition is counted once.

\textbf{Step 5: repeat corpus coding.} A coder is a separate model run, not a human participant. Each receives the full codebook and batches of fixed units, returning all applicable subcodes and ordinal levels. Units may receive multiple codes or an empty set. Requested and effective settings, batch size, retries and treatment of omitted outputs must be recorded. Separate calls provide repeated applications but do not create independent interpretive perspectives in the sense of separately trained human coders.

\textbf{Step 6: assess representation and agreement.} Compare codebook size, scale declarations and coverage alongside within-model and between-model reliability. Code count alone can reward fragmentation; coverage alone can reward broad categories. This profile exposes trade-offs without ranking substantive quality. The difference between own- and shared-codebook reliability is descriptive: because size, wording and category boundaries also change, it does not isolate self-preference.

\textbf{Step 7: measure conditional dispersion.} Aggregate assignments only within a specified category system. The shared reference provides a common vocabulary for describing concentration or dispersion. State whether residual codes, repeated coders and multiple assignments per unit are included. High entropy means that assignments are widely distributed under that convention; it does not prove saturation, validity or richer evidence.

\textbf{Step 8: interpret and document.} Connect differences to the intended substantive claims. A distinctive codebook requires inspection of definitions and supporting material; high within-model agreement with lower between-model agreement warrants caution about treating one model as definitive. Human adjudication can assess the significance of disagreements and missed distinctions. This application reports the diagnostics and implications; human validation remains unperformed and is not claimed as an outcome.

\subsection{Metric specification}
Codebook \emph{size} is the number of substantive subcodes. \emph{Ordinal share} is the proportion of all codes declared ordinal, counting the residual as nominal. Scale declarations record model judgements, not independently established measurement properties. \emph{Coverage} is the fraction of all units receiving at least one code, including the residual. It therefore does not measure recall against human judgements of relevance.

Coding is multi-label. A coder's value for a unit is a set of assigned identifiers. We use Krippendorff's $\alpha$ with MASI distance, which distinguishes identical, nested, partially overlapping and disjoint sets \parencite{passonneau2006}. MASI distance is one minus Jaccard similarity multiplied by an overlap weight: 1 for identical sets, $2/3$ for strict containment, $1/3$ for partial overlap and 0 for disjoint sets. Two empty sets have distance zero. Reliability is
\begin{equation}
\alpha=1-\frac{D_o}{D_e},
\end{equation}
where $D_o$ is observed disagreement and $D_e$ is disagreement expected from the theme's full distribution of coder values. We calculate $D_e$ exactly from all units. Within-model $\alpha$ compares three runs of one model under one codebook. Between-model $\alpha$ compares the first run of each model on the shared codebook, separately by theme. Summaries are unweighted theme means, so large deployments do not dominate them.

The original analysis also computed unit-bootstrap intervals for between-model agreement (1,000 resamples, fixed seed), holding $D_e$ at its full-sample value. Units are nested within interviews; resampling them independently does not account for clustering, and fixing $D_e$ omits uncertainty in the chance term. We therefore present descriptive point estimates and do not interpret interval separation as a significance test. Conventional $\alpha$ thresholds provide context but cannot certify validity for repeated calls to related models \parencite{krippendorff2019}.

For qualitative variance, $p_c$ is a code's share of all assignments on a theme's shared codebook. The application pools all 24 coders (eight models, three runs), counts each assignment once and includes the residual. A multi-coded unit contributes more assignments than a singly coded unit; empty units contribute none. We report
\begin{equation}
H=-\sum_c p_c\log_2 p_c,\quad N_{\mathrm{eff}}=2^H,\quad H'=\frac{H}{\log_2 k},\quad D=\max_c p_c,
\end{equation}
where $k$ is the number of distinct codes used. $N_{\mathrm{eff}}$ expresses entropy as the effective number of equally frequent codes \parencite{hill1973diversity,jost2006entropy}; $H'$ is normalised entropy, and $D$ the largest assignment share. These describe a pooled coded representation, not respondent-level diversity. They may reflect both variation in material and coder disagreement. Ordinal levels were recorded but are not analysed here.

\section{Empirical Application}
\subsection{Corpus and model panel}
The corpus comprises interviews conducted on one conversational-survey platform in one organisation. Within each deployment, an LLM interviewer follows a semi-structured protocol and adapts follow-up questions to the respondent. Six deployments with distinct protocols and audiences form the themes in Table~\ref{tab:corpus}. Five are in English; Employee Engagement is coded in German to avoid adding translation as another model-dependent operation.

Of 1,542 raw transcript records, 893 pass cleaning: they parse, contain at least two respondent turns and 30 respondent words, and are not duplicates. We then exclude two deployments with 20 and two interviews, demonstration and test interviews, and non-German Employee Engagement interviews. The final corpus contains 668 interviews. Respondent turns of up to 80 words remain intact; longer turns are split at sentence boundaries, producing 13,166 units. Almost all units contain at most 80 words. Models receive respondent text without interviewer questions, a shared restriction that can remove relevant context.

\begin{table}[!htbp]
\centering\tablefont
\begin{tabularx}{\textwidth}{Ylrr}
\toprule
\textbf{Theme} & \textbf{Language} & \textbf{Interviews} & \textbf{Units}\\
\midrule
AI Assessment & English & 179 & 4,266\\
AI Training & English & 95 & 1,821\\
AI Usage and Literacy & English & 61 & 1,677\\
Teaming Check-ins & English & 205 & 2,788\\
Employee Engagement & German & 96 & 2,222\\
Minority Empowerment & English & 32 & 392\\
\midrule
Total & & 668 & 13,166\\
\bottomrule
\end{tabularx}
\caption{Study corpus by theme. Units are fixed before any model comparison.}
\label{tab:corpus}
\end{table}

The panel combines five vendors with a within-vendor comparison: Opus 4.8, Sonnet 5, Haiku 4.5 and Fable 5 from Anthropic; GPT-4.1 from OpenAI; Gemini 2.5 Pro from Google; Grok 4.5 from xAI; and DeepSeek v4-pro. These are the model labels used in the study records. Sizes, generations, access routes and sampling configurations are not matched. Results characterise the recorded configurations rather than causal vendor effects or a current purchasing recommendation.

\subsection{Instantiation of the eight steps}
For each theme, all models receive the same randomly selected 15 interviews with a fixed selection seed. The creation prompt (Appendix~\ref{app:prompts}) requests at most six categories and five substantive subcodes per category. It asks for the fewest well-separated codes that cover the sample, prohibits padding and additional catch-all categories, and requires the components in Step 3. The pipeline adds \texttt{OTHER\_RELEVANT}, defined as relevant content that fits no substantive code. One codebook per model per theme yields 48 instruments; creation is not repeated.

Opus is the shared-reference author, selected before codebook creation. All models code all units using their own and the theme's Opus codebook, with three runs per condition. This gives $6\times3\times(2\times8-1)=270$ coding runs. Metadata identify the codebook creator, so application is not blind to provenance. Each call contains the theme name, full JSON codebook and 60 units; DeepSeek uses 20 and Gemini 30 to reduce truncation. The prompt requires one response per unit, all applicable subcodes, and one allowed level for each ordinal code. It permits an empty list when no code applies. Invalid identifiers are discarded; omitted units in otherwise valid responses are treated as uncoded. Unparseable batches are retried up to four total attempts. Empty assignments can thus conflate a coding decision with an output failure.

Creation requests temperature 0 with seed 7000; coding requests temperature 1 with distinct seeds per coder. Actual controls depend on the route (Table~\ref{tab:settings}). OpenAI, xAI and DeepSeek receive temperature and seed; Google receives temperature; Anthropic routes use default sampling. Creation otherwise uses the coding routes except that five Sonnet codebooks are created through the subscription command-line interface (CLI) and one through the API, and Gemini uses default reasoning for creation. Sonnet creation token counts suggest active reasoning. Effective decoding settings are therefore not identical across models.

\begin{table}[!htbp]
\tablefont
\begin{tabularx}{\textwidth}{L{2.4cm}L{4.1cm}L{2.1cm}Y}
\toprule
\textbf{Model} & \textbf{Recorded identifier} & \textbf{Coding route} & \textbf{Sampling and reasoning}\\
\midrule
Opus 4.8 & \texttt{claude-opus-4-8} & Subscription CLI & Default; zero reasoning requested, likely active.\\
Sonnet 5 & \texttt{claude-sonnet-5} & Anthropic API & Default; reasoning disabled; one CLI run.\\
Haiku 4.5 & \texttt{claude-haiku-4-5-}\allowbreak\texttt{20251001} & Subscription CLI & Default; zero requested, little apparent reasoning.\\
Fable 5 & \texttt{claude-fable-5} & Anthropic API & Default sampling and reasoning.\\
GPT-4.1 & \texttt{gpt-4.1} & OpenAI API & Temperature 1 and seed; no reasoning mode.\\
Gemini 2.5 Pro & \texttt{gemini-2.5-pro} & Google API & Temperature 1; reasoning budget 128 tokens.\\
Grok 4.5 & \texttt{grok-4.5} & xAI API & Temperature 1 and seed; default reasoning.\\
DeepSeek v4-pro & \texttt{deepseek-v4-pro} & DeepSeek API & Temperature 1 and seed; default reasoning.\\
\bottomrule
\end{tabularx}
\tabnote{Identifiers are aliases unless dated; GPT-4.1 resolved to \texttt{gpt-4.1-2025-04-14}. Sonnet's CLI exception is the first shared-codebook run for Minority Empowerment. Requested seeds do not establish deterministic execution.}
\caption{Access routes and settings in the empirical application.}
\label{tab:settings}
\end{table}

A control study uses 60 Teaming Check-ins units and three runs per condition to compare Sonnet with reasoning on/off and Gemini with default/reduced reasoning. Sonnet's within-model $\alpha$ is .840 with reasoning and .838 without; Gemini's is .760 with default reasoning and .815 with the reduced budget. Assignments also change. This motivates documenting configurations; it does not remove the main comparison's decoding confound or assess reasoning during creation.

\section{Results}
\subsection{Generated instruments differ in granularity and scale declarations}
Models produce different category systems from the same samples and instructions (Table~\ref{tab:results}). GPT-4.1 averages 15.0 substantive subcodes, compared with 24.8 for Fable and 25.0 for DeepSeek. This changes the distinctions available to subsequent analysis; it does not rank quality. Higher granularity does not consistently produce higher coverage: DeepSeek and Fable code .663 and .700 of units, compared with .741 and .745 for Opus and Sonnet. GPT-4.1 has the lowest coverage (.606).

The workload area in Teaming Check-ins illustrates the consequence. Opus separates overall sustainability, working hours, workload drivers, protected time and recovery, and stress or energy. GPT-4.1 uses three subcodes: workload intensity and sustainability, task prioritisation and allocation, and fragmentation and task switching. Analyses based on these instruments can directly distinguish different features of the same material. The example makes that choice visible for review without establishing which grouping is better.

Scale declarations also vary. On Employee Engagement, Haiku declares 15 of its 16 substantive subcodes ordinal, while DeepSeek declares none of its 27 ordinal. On Teaming Check-ins, the ordinal share ranges from .09 for DeepSeek to .88 for Gemini. Accepting these declarations automatically would import model-dependent decisions about which phenomena admit ordered measurement. The audit exposes this dependence before an ordinal summary is interpreted.

\begin{table}[!htbp]
\centering\tablefont
\begin{tabularx}{\textwidth}{Yrrrrrr}
\toprule
& \multicolumn{3}{c}{\textbf{Own-codebook representation}} & \multicolumn{3}{c}{\textbf{Within-model reliability}}\\
\cmidrule(lr){2-4}\cmidrule(lr){5-7}
\textbf{Model} & \textbf{Subcodes} & \textbf{Ordinal share} & \textbf{Coverage} & \textbf{Own} & \textbf{Shared} & \textbf{Difference}\\
\midrule
Opus 4.8 & 22.2 & .29--.74 & .741 & .861 & .861 & ---\\
Sonnet 5 & 22.0 & .08--.81 & .745 & .861 & .843 & +.018\\
Haiku 4.5 & 17.5 & .21--.88 & .670 & .804 & .819 & -.015\\
Fable 5 & 24.8 & .13--.31 & .700 & .917 & .909 & +.008\\
GPT-4.1 & 15.0 & .00--.53 & .606 & .837 & .838 & -.001\\
Gemini 2.5 Pro & 17.3 & .05--.88 & .657 & .798 & .821 & -.023\\
Grok 4.5 & 23.5 & .00--.63 & .681 & .831 & .856 & -.024\\
DeepSeek v4-pro & 25.0 & .00--.58 & .663 & .764 & .782 & -.018\\
\bottomrule
\end{tabularx}
\tabnote{Subcodes: theme mean, excluding residual. Ordinal share: theme range, residual nominal. Coverage: mean coded-unit share across themes and runs. Reliability: three-run MASI $\alpha$, averaged across themes. Difference: own minus shared before rounding; Opus's conditions coincide.}
\caption{Category-system diagnostics and application reliability by model.}
\label{tab:results}
\end{table}

\subsection{Repeated application is more consistent than cross-model application}
\begin{figure}[H]
\centering\includegraphics[width=0.95\textwidth]{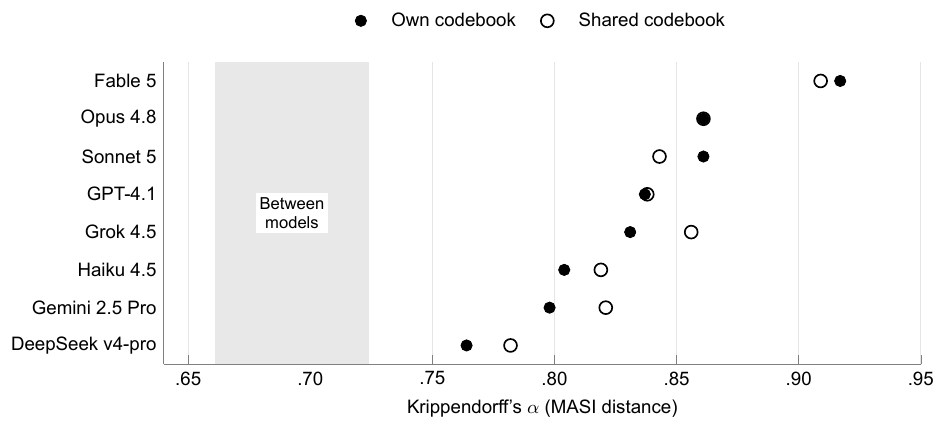}
\caption{Within-model reliability on own and shared codebooks (unweighted theme means). Shading shows the between-model range across themes, not a confidence interval.}
\label{fig:reliability}
\end{figure}

Mean within-model $\alpha$ ranges from .764 to .917 on own codebooks and from .782 to .909 on the shared reference. Between-model agreement is lower (.661--.724 across themes; Table~\ref{tab:themes}). Figure~\ref{fig:reliability} shows this gap: a model can code consistently across runs while other models interpret the same categories differently.

The difference in mean $\alpha$ (own minus shared) ranges from -.024 to +.018, averaging about -.008 across the seven non-reference models. Five of seven are slightly more consistent on the shared codebook. Own codebooks thus offer no consistent reliability advantage. This does not rule out self-preference: the instrument changes and its creator is visible.

Two empty sets count as agreement. Models leave roughly one quarter to two fifths of units uncoded on own codebooks, including both explicit non-assignment and omitted outputs. Reliability therefore includes uncoded material and must be read alongside coverage.

\subsection{Vendor grouping explains only a limited part of the disagreement}
The four Anthropic models agree somewhat more with one another than the four models from other vendors do (Table~\ref{tab:themes}). Mean agreement is about .71 compared with .68 for the other group. Both remain below the within-model means. This comparison is asymmetric: one group shares a vendor, whereas the other contains four vendors, and the shared codebook was written by an Anthropic model. It does not identify a vendor effect or show that different vendors guarantee independent judgements.

\begin{table}[!htbp]
\centering\tablefont
\begin{tabularx}{\textwidth}{Yrrrrrrr}
\toprule
& \multicolumn{3}{c}{\textbf{Between-model $\alpha$}} & \multicolumn{4}{c}{\textbf{Conditional code dispersion}}\\
\cmidrule(lr){2-4}\cmidrule(lr){5-8}
\textbf{Theme} & \textbf{All 8} & \textbf{Anthropic} & \textbf{Other} & $\boldsymbol{k}$ & $\boldsymbol{N_{\mathrm{eff}}}$ & $\boldsymbol{H'}$ & $\boldsymbol{D}$\\
\midrule
AI Assessment & .677 & .705 & .650 & 19 & 16.0 & .942 & .120\\
AI Training & .707 & .720 & .693 & 21 & 15.4 & .898 & .164\\
AI Usage and Literacy & .724 & .739 & .711 & 23 & 20.0 & .955 & .094\\
Teaming Check-ins & .664 & .686 & .653 & 23 & 19.3 & .943 & .099\\
Employee Engagement & .661 & .681 & .650 & 26 & 22.6 & .958 & .103\\
Minority Empowerment & .722 & .737 & .707 & 27 & 22.6 & .946 & .098\\
\bottomrule
\end{tabularx}
\tabnote{All values use the shared codebook. Between-model $\alpha$ uses one run per model; each vendor group has four models. Dispersion pools all 24 coders' assignments, including residual codes. $k$: distinct codes used; $N_{\mathrm{eff}}$: effective count; $H'$: normalised entropy; $D$: largest code share.}
\caption{Agreement and codebook-conditional qualitative variance by theme.}
\label{tab:themes}
\end{table}

\subsection{Code use is dispersed, conditional on the reference instrument}
Normalised entropy is high: .898 for AI Training and .942--.958 for the other themes. Between 15.4 and 22.6 codes are effectively in use, out of 19--27 distinct codes. The most frequent code accounts for at most .164 of assignments. No single code therefore dominates the pooled representation. The effective number also closely follows reference-codebook size across these six themes, underscoring its dependence on the instrument.

Employee Engagement has the highest normalised entropy and lowest between-model agreement (.661). Broad code use and agreement describe different properties. With one German theme, language cannot be separated from topic and deployment. High entropy does not show that interviewing has reached saturation: repeated coder assignments, granularity and mixing of coder interpretations contribute to the distribution.

Cost is an operational consideration rather than a validity criterion. At logged token counts and study-time list prices, coding 1,000 units once ranges from approximately US\$0.54 for GPT-4.1 to US\$8.96 for Fable. Subscription-route costs are estimates; incomplete accounting of separately reported reasoning tokens makes some costs lower bounds. These configuration-specific values do not establish the cost-effectiveness of valid qualitative inference.

\section{Discussion}
\subsection{Contribution of the framework}
QVAF connects a choice made before coding---which distinctions the model creates---to the resulting coded representation. An application-only benchmark cannot observe those initial choices. A comparison of generated themes alone cannot show how consistently the instruments work on the full corpus. Combining both with a shared-reference condition enables a more specific diagnosis: structural differences appear in generated codebooks, while the reference reveals disagreement remaining when categories are fixed.

The application demonstrates why these diagnostics belong together. More subcodes do not consistently yield more coverage; repeated calls can be highly consistent while models disagree on the same instrument; and high entropy can coexist with lower agreement. None of these quantities substitutes for the others. The contribution is a comparison and reporting logic that prevents a favourable result on one dimension from being interpreted as overall quality.

The framework complements LLM-assisted grounded theory. GGT addresses how researchers move from source material towards theory \parencite{schmitt2026}; QVAF asks how the category systems and coded evidence used along that path depend on a model configuration. Researchers can insert the audit before committing to an instrument or use it to inspect an existing analysis. Human interpretation is required to decide whether a model-specific category is informative, overfragmented or misleading. The present results identify targets for that judgement without claiming to resolve them.

\subsection{Implications for IS research and organisational use}
For IS researchers using AI-generated categories, the practical output is an inspectable instrument and a sensitivity account. An application should report the common creation sample, generation specification, codebook structure, unitisation, repeated within-model reliability, cross-model agreement under a stated reference and the code-use denominator. Definitions and examples remain essential: a reliable label can encode a poor conceptual distinction. Where substantive conclusions depend on particular categories, researchers should inspect underlying material and obtain independent human assessment before treating the audit as validity evidence.

For organisations analysing interviews or open feedback at scale, consistent processing by one model does not ensure that another would produce the same interpretation. Standardising on a provider does not itself test representational dependence. Comparing at least two materially different model configurations can reveal consequential differences; this study establishes neither an optimal panel nor independent perspectives across vendors. Codebook provenance, provider settings and hosted-model changes should accompany results so later comparisons remain interpretable.

\subsection{Limitations and scope}
First, no human reference analysis or independent validity assessment is available. The framework diagnoses differences but cannot identify the most appropriate codebook. Human review should assess category content and assignments, including cases left empty. Demonstrating improvement in subsequent theory or organisational decisions requires another evaluation.

Second, each model creates one codebook per theme from one sample and prompt. Structural differences cannot be separated from stochastic creation variation, sample sensitivity or prompt interactions. Repeated creation, alternative samples and prompt variants are priority robustness checks, not completed analyses. The eight-step procedure is portable in form; its effectiveness across settings remains to be evaluated.

Third, the reference is one model's instrument. Multiple references or a fully crossed design would test whether agreement differences persist. Sampling, reasoning, routes and batch sizes also vary. The 60-unit reasoning control does not resolve those confounds. Since creator metadata are visible, provenance may also affect coding.

Fourth, the corpus comes from one platform and organisation, is mostly English, and consists of AI-led interviews. Human-led interviews, other domains and languages may yield different findings. The smallest theme has 32 interviews. Removing interviewer questions can obscure meaning, and the 80-word segmentation threshold has not been sensitivity-tested.

Finally, between-model agreement uses one run per model, and the unit-bootstrap procedure omits within-interview clustering. Empty-set agreement and omitted outputs affect reliability. Pooled dispersion weights assignments rather than respondents and combines material variation with coder variation. It cannot support causal claims about vendors, inferences about population diversity or conclusions about theoretical saturation. Scale does not resolve these interpretive limits.

\section{Conclusion}
The Qualitative Variance Audit Framework organises comparison of LLM-generated category systems into eight steps, from a fixed corpus and common elicitation inputs to repeated application, agreement diagnostics and codebook-conditional dispersion. It provides an explicit audit of how model choice shapes qualitative representation. Across 668 interviews, eight models generate instruments of different granularity and scale type, and high within-model consistency coexists with substantially lower cross-model agreement on a shared reference. The framework complements grounded-theory and IS taxonomy work by exposing these dependencies before coded outputs support substantive claims. Establishing their validity remains a task for researchers and independent evaluation.

\section*{Research Transparency Statements}
Ethics approval covers the complete study corpus. Respondents were informed that the interviewer was an AI agent and that responses would be used for research. The data owner authorised research use and processing by external model providers. Transcripts were processed without prior anonymisation through the recorded routes. The paper contains no respondent quotations or personal identifiers. The corpus is proprietary; release of transcripts or derived material requires data-owner approval and appropriate anonymisation.

The research team developed the research questions and conceptual framing and conducted the literature review. Generative AI was used to generate the codebooks and code the interview corpus, as described in the methods. AI tools also supported technical implementation and preparation of the manuscript, including programming assistance, language editing, bibliographic checks and formatting. The research team made the substantive methodological and interpretive decisions and retains responsibility for the analysis, source selection and final manuscript.

The study received no funding. The authors have a professional relationship with the organisation owning the data. There are no further competing interests.

The paper specifies the design, metrics, prompt requirements and model configurations. Transcripts and derived study artefacts are not publicly released. Access or release is subject to data-owner authorisation. Reported aggregates use archived analysis outputs; not all original run-level artefacts remain available for independent recomputation.

\printbibliography[title={References}]

@inproceedings{xiao2023deductive,
  author    = {Xiao, Ziang and Yuan, Xingdi and Liao, Q. Vera and Abdelghani, Rania and Oudeyer, Pierre-Yves},
  title     = {Supporting qualitative analysis with large language models: Combining codebook with {GPT-3} for deductive coding},
  booktitle = {Companion Proceedings of the 28th International Conference on Intelligent User Interfaces (IUI '23 Companion)},
  pages = {75--78},
  publisher = {ACM},
  year      = {2023},
  doi       = {10.1145/3581754.3584136}
}

@article{depaoli2024inductive,
  author  = {De Paoli, Stefano},
  title   = {Performing an inductive thematic analysis of semi-structured interviews with a large language model: An exploration and provocation on the limits of the approach},
  journal = {Social Science Computer Review},
  year    = {2024},
  volume  = {42},
  number  = {4},
  pages   = {997--1019},
  doi     = {10.1177/08944393231220483}
}

@article{ashwin2025bias,
  author  = {Ashwin, Julian and Chhabra, Aditya and Rao, Vijayendra},
  title   = {Using large language models for qualitative analysis can introduce serious bias},
  journal = {Sociological Methods \& Research},
  year    = {2026},
  volume = {55},
  number = {3},
  pages = {795--839},
  doi     = {10.1177/00491241251338246},
}

@article{braun2019reflecting,
  author  = {Braun, Virginia and Clarke, Victoria},
  year    = {2019},
  title   = {Reflecting on reflexive thematic analysis},
  journal = {Qualitative Research in Sport, Exercise and Health},
  volume  = {11},
  number  = {4},
  pages   = {589--597},
  doi     = {10.1080/2159676X.2019.1628806}
}

@article{macqueen1998codebook,
  author  = {MacQueen, Kathleen M. and McLellan, Eleanor and Kay, Kelly and Milstein, Bobby},
  year    = {1998},
  title   = {Codebook development for team-based qualitative analysis},
  journal = {Cultural Anthropology Methods},
  volume  = {10},
  number  = {2},
  pages   = {31--36},
  doi     = {10.1177/1525822X980100020301}
}

@article{hennink2022sample,
  author  = {Hennink, Monique and Kaiser, Bonnie N.},
  year    = {2022},
  title   = {Sample sizes for saturation in qualitative research: {A} systematic review of empirical tests},
  journal = {Social Science \& Medicine},
  volume  = {292},
  doi     = {10.1016/j.socscimed.2021.114523},
  eid = {114523}
}

@article{hayes2007answering,
  author  = {Hayes, Andrew F. and Krippendorff, Klaus},
  year    = {2007},
  title   = {Answering the call for a standard reliability measure for coding data},
  journal = {Communication Methods and Measures},
  volume  = {1},
  number  = {1},
  pages   = {77--89},
  doi     = {10.1080/19312450709336664}
}

@article{krippendorff2004reliability,
  author  = {Krippendorff, Klaus},
  year    = {2004},
  title   = {Reliability in content analysis: {Some} common misconceptions and recommendations},
  journal = {Human Communication Research},
  volume  = {30},
  number  = {3},
  pages   = {411--433},
  doi     = {10.1111/j.1468-2958.2004.tb00738.x}
}

@article{oconnor2020intercoder,
  author  = {O'Connor, Cliodhna and Joffe, Helene},
  year    = {2020},
  title   = {Intercoder reliability in qualitative research: {Debates} and practical guidelines},
  journal = {International Journal of Qualitative Methods},
  volume  = {19},
  doi     = {10.1177/1609406919899220},
  eid = {1609406919899220}
}

@article{campbell2013coding,
  author  = {Campbell, John L. and Quincy, Charles and Osserman, Jordan and Pedersen, Ove K.},
  year    = {2013},
  title   = {Coding in-depth semistructured interviews: {Problems} of unitization and intercoder reliability and agreement},
  journal = {Sociological Methods \& Research},
  volume  = {42},
  number  = {3},
  pages   = {294--320},
  doi     = {10.1177/0049124113500475}
}

@article{hill1973diversity,
  author  = {Hill, Mark O.},
  year    = {1973},
  title   = {Diversity and evenness: {A} unifying notation and its consequences},
  journal = {Ecology},
  volume  = {54},
  number  = {2},
  pages   = {427--432},
  doi     = {10.2307/1934352}
}

@article{jost2006entropy,
  author  = {Jost, Lou},
  year    = {2006},
  title   = {Entropy and diversity},
  journal = {Oikos},
  volume  = {113},
  number  = {2},
  pages   = {363--375},
  doi     = {10.1111/j.2006.0030-1299.14714.x}
}

@inproceedings{pham2024topicgpt,
  author    = {Pham, Chau Minh and Hoyle, Alexander and Sun, Simeng and Resnik, Philip and Iyyer, Mohit},
  title     = {{TopicGPT}: A prompt-based topic modeling framework},
  booktitle = {Proceedings of the 2024 Conference of the North American Chapter of the Association for Computational Linguistics: Human Language Technologies (NAACL 2024), Long Papers},
  year      = {2024},
  pages     = {2956--2984},
  publisher = {Association for Computational Linguistics},
  doi = {10.18653/v1/2024.naacl-long.164},
  url       = {https://aclanthology.org/2024.naacl-long.164/}
}

@article{nickerson2013method,
  author  = {Nickerson, Robert C. and Varshney, Upkar and Muntermann, Jan},
  title   = {A Method for Taxonomy Development and its Application in Information Systems},
  journal = {European Journal of Information Systems},
  year    = {2013},
  volume  = {22},
  number  = {3},
  pages   = {336--359},
  doi     = {10.1057/ejis.2012.26}
}

@inproceedings{passonneau2006,
 author={Passonneau, Rebecca},
 title={Measuring agreement on set-valued items ({MASI}) for semantic and pragmatic annotation},
 booktitle={Proceedings of the Fifth International Conference on Language Resources and Evaluation},
 year={2006}, pages={831--836}, publisher={European Language Resources Association},
 url={https://aclanthology.org/L06-1392/}
}

@book{krippendorff2019,
 author={Krippendorff, Klaus}, year={2019},
 title={Content analysis: An introduction to its methodology}, edition={4}, publisher={SAGE}
}

@article{blohm2025,
  author = {Blohm, Ivo and Miranda, Shaila and Ho, Shuk Ying and Leimeister, Jan Marco},
  title = {Next-generation {IS} research methods---towards a better understanding of complex and dynamic phenomena \ldots{} and generative {AI} as the elephant in the room},
  journal = {Journal of Information Technology},
  year = {2025},
  volume = {40},
  number = {2},
  pages = {102--121},
  doi = {10.1177/02683962251340699},
  url = {https://doi.org/10.1177/02683962251340699}
}

@misc{arlinghaus2024,
  author = {Arlinghaus, Clarissa Sabrina and Wulff, Charlotte and Maier, G{\"u}nter W.},
  title = {Inductive Coding with {ChatGPT}---An Evaluation of Different {GPT} Models Clustering Qualitative Data into Categories},
  year = {2024},
  howpublished = {OSF Preprints},
  note = {Preprint},
  doi = {10.31219/osf.io/gpnye},
  url = {https://doi.org/10.31219/osf.io/gpnye}
}

@article{dorfel2026,
  author = {D{\"o}rfel, Leoni and Ammoneit, Rieke},
  title = {Evaluation of Inductive Coding with {LLMs}},
  journal = {Education Sciences},
  year = {2026},
  volume = {16},
  number = {8},
  doi = {10.3390/educsci16081314},
  url = {https://doi.org/10.3390/educsci16081314},
  eid = {1314}
}

@article{schmitt2026,
  author = {Schmitt, Bernd and Hao, Shuyi and Pham, Michel Tuan and Hofstetter, Reto},
  title = {Generative grounded theory ({GGT}): Inductive theory building in the age of generative {AI}},
  journal = {Journal of Consumer Psychology},
  year = {2026},
  volume = {36},
  number = {4},
  pages = {500--520},
  doi = {10.1002/jcpy.70036},
  url = {https://doi.org/10.1002/jcpy.70036}
}

@inproceedings{sinha2024,
  author = {Sinha, Ravi and Solola, Idris and Nguyen, Ha and Swanson, Hillary and Lawrence, LuEttaMae},
  title = {The Role of Generative {AI} in Qualitative Research: {GPT-4}'s Contributions to a Grounded Theory Analysis},
  booktitle = {Proceedings of the 2024 Symposium on Learning, Design and Technology},
  year = {2024},
  pages = {17--25},
  publisher = {Association for Computing Machinery},
  address = {New York, NY},
  doi = {10.1145/3663433.3663456},
  url = {https://doi.org/10.1145/3663433.3663456}
}

@article{bennis2025,
  author = {Bennis, Issam and Mouwafaq, Safwane},
  title = {Advancing {AI}-driven thematic analysis in qualitative research: a comparative study of nine generative models on {Cutaneous Leishmaniasis} data},
  journal = {BMC Medical Informatics and Decision Making},
  year = {2025},
  volume = {25},
  doi = {10.1186/s12911-025-02961-5},
  url = {https://doi.org/10.1186/s12911-025-02961-5},
  eid = {124}
}

@article{berente2019,
  author = {Berente, Nicholas and Seidel, Stefan and Safadi, Hani},
  title = {Research Commentary---Data-Driven Computationally Intensive Theory Development},
  journal = {Information Systems Research},
  year = {2019},
  volume = {30},
  number = {1},
  pages = {50--64},
  doi = {10.1287/isre.2018.0774},
  url = {https://doi.org/10.1287/isre.2018.0774}
}

@article{tai2024,
 author={Tai, Robert H. and Bentley, Lillian R. and Xia, Xin and Sitt, Jason M. and Fankhauser, Sarah C. and Chicas-Mosier, Ana M. and Monteith, Barnas G.},
 title={An examination of the use of large language models to aid analysis of textual data},
 journal={International Journal of Qualitative Methods}, year={2024},volume={23},
 doi={10.1177/16094069241231168},
  eid = {16094069241231168}
}
\appendix
\section{Research Prompts}
\label{app:prompts}
The system and user prompts below specify the common elicitation and coding tasks. Wording is retained from the study templates; Markdown emphasis is typeset and the JSON schemas are compacted. Placeholders are filled with the theme name, $K=15$, a maximum of six categories and five subcodes per category, the sampled respondent turns, the complete codebook, and the current unit batch. The prompts contain no respondent data. Operational settings and departures from requested controls are documented in Section~4.2.

\subsection{Codebook creation}
\textbf{System}

You are an expert qualitative researcher building a codebook for thematic analysis. You produce codebooks whose categories are mutually exclusive where possible, collectively cover the material, and are defined precisely enough for a second coder to apply them consistently without you present. You design the category system only --- you do not code the data in this task.

\textbf{User}

You are given a random sample of \{K\} interview transcripts from a single study theme:\\ "\{THEME\_NAME\}". Each interview was AI-led; only the respondent turns are shown, because only they carry content.

The sampled transcripts follow. Each line is one respondent-turn unit, rendered as \texttt{(unit\_id) text} and grouped under its interview heading:

\{SAMPLE\_TRANSCRIPTS\}

Build a \textbf{two-level codebook} for this theme, grounded strictly in what appears in the sample:

\begin{itemize}
\item \textbf{Level 1 --- top categories:} at most \{MAX\_TOP\}. Broad, well-separated topic areas.
\item \textbf{Level 2 --- subcodes:} at most \{MAX\_SUB\} per category. Each subcode is one applyable code.
\end{itemize}
A reserved residual subcode called \texttt{OTHER\_RELEVANT} --- "content that is relevant to the theme but does not fit any other code" --- \textbf{already exists} and will be added to your codebook automatically. Do \textbf{not} create it yourself, and do \textbf{not} add any other "miscellaneous", "other", or "general" catch-all code. Every category and subcode you write must name a specific, substantive topic.

For \textbf{every subcode} provide, exactly:

\begin{itemize}
\item \texttt{label} --- short, distinct, human-readable.
\item \texttt{definition} --- 1--2 sentences.
\item \texttt{inclusion} --- what content this code covers (when to apply it).
\item \texttt{exclusion} --- near-miss content this code does NOT cover (to separate it from its neighbours).
\item \texttt{example} --- one short paraphrased illustration (no names, no identifying detail).
\item \texttt{scale\_type} --- \textbf{required.} \texttt{"ordinal"} if a unit's instances of this subcode can be meaningfully
ordered on one dimension (e.g., low$\rightarrow$high satisfaction, never$\rightarrow$always); otherwise \texttt{"nominal"}.
\item \texttt{ordered\_levels} --- \textbf{required iff \texttt{scale\_type} is \texttt{"ordinal"}}: an array of 3--7 ordered level labels
from lowest to highest (e.g. \texttt{["low","medium","high"]}). Set to \texttt{null} when \texttt{scale\_type} is
\texttt{"nominal"}.
\end{itemize}
Rules:

\begin{itemize}
\item Cover as much of the sampled material as you can with as \textbf{few, well-separated} codes as the caps
allow. Do not pad to the cap. Do not create near-duplicate codes.
\item Every code must be groundable in the sample --- no speculative categories, no codes an unseen dataset
"might" need.
\item Prefer specific codes over broad ones: a code so broad it would absorb most units is a defect, not a
win.
\end{itemize}
\noindent\begin{minipage}{\linewidth}
Return \textbf{only} JSON matching this schema, and nothing else (no prose, no markdown fences):

\begin{Verbatim}[breaklines=true,samepage=true,fontsize=\normalsize]
{"theme": "string", "categories": [{"label": "string", "subcodes": [{
  "label": "string", "definition": "string", "inclusion": "string",
  "exclusion": "string", "example": "string",
  "scale_type": "ordinal | nominal",
  "ordered_levels": ["low", "...", "high"]}]}]}
\end{Verbatim}
\end{minipage}\par

\subsection{Codebook application}
\textbf{System}

You are a careful qualitative coder. You apply a given codebook to text units exactly as defined, using each code's inclusion and exclusion criteria. You never invent codes, never rename them, and never add commentary. You return only valid JSON.

\textbf{User}

Theme: "\{THEME\_NAME\}". Apply the codebook below to each unit.

Codebook (use the \texttt{code\_id} values exactly as given): \{CODEBOOK\_JSON\}

Units to code: \{UNITS\}

Coding rules:

\begin{itemize}
\item For each unit, assign \textbf{all} applicable subcodes (multi-label). A unit may receive zero, one, or
several \texttt{code\_id}s.
\item Assign a subcode only when the unit's text satisfies that code's \texttt{inclusion} and is not ruled out by
its \texttt{exclusion}.
\item Use \texttt{OTHER\_RELEVANT} only for content that is clearly relevant to the theme but fits no substantive
subcode. Do not use it for off-topic content.
\item If a unit has no relevant content, return an \textbf{empty} \texttt{codes} list for it (do not drop the unit).
\item For every assigned subcode whose \texttt{scale\_type} is \texttt{"ordinal"}, also report its level in \texttt{levels},
choosing exactly one value from that subcode's \texttt{ordered\_levels}. Omit nominal codes from \texttt{levels}.
(If capturing levels is disabled, omit the \texttt{levels} field entirely.)
\item Return \textbf{exactly one} entry for \textbf{every} \texttt{unit\_id} given in this batch --- no missing ids, no extra
ids, no duplicates.
\end{itemize}
\noindent\begin{minipage}{\linewidth}
Return \textbf{only} JSON matching this schema, and nothing else (no prose, no markdown fences):

\begin{Verbatim}[breaklines=true,samepage=true,fontsize=\normalsize]
{"codings": [{"unit_id": "string", "codes": ["<code_id>", "..."],
  "levels": {"<ordinal code_id>": "<one of its ordered_levels>"}}]}
\end{Verbatim}
\end{minipage}\par
Use \texttt{code\_id} values exactly as written in the codebook. Do not add any keys beyond those shown.

\end{document}